\documentclass[runningheads]{llncs}

\usepackage{eccv}

\usepackage{eccvabbrv}

\usepackage{graphicx}
\usepackage{booktabs}
\usepackage{caption}
\usepackage[accsupp]{axessibility}  

\usepackage{graphicx}
\usepackage{floatrow}
\usepackage{multirow}
\usepackage{xcolor}
\usepackage{colortbl}
\usepackage{amssymb}
\usepackage{pifont}
\newcommand{\cmark}{\ding{51}}%
\newcommand{\xmark}{\ding{55}}%
\usepackage{hyperref}

\usepackage{orcidlink}
\newcommand{\swap}[1]{\textcolor{black}{#1}}
\def\modelone{{MoCA}}

\begin{document}

\title{Unsupervised Audio-Visual Segmentation with Modality Alignment} 

\titlerunning{UAVS}

\author{Swapnil Bhosale\inst{1} \and
Haosen Yang\inst{1} \and
Diptesh Kanojia\inst{1} \and
Jiankang Deng\inst{2} \and
Xiatian Zhu\inst{1}}

\authorrunning{Bhosale et al.}

\institute{University of Surrey, UK \and
Imperial College London, UK}

\maketitle
\begin{abstract}

Audio-Visual Segmentation (AVS) aims to identify, at the pixel level, the object in a visual scene that produces a given sound. 
Current AVS methods rely on costly fine-grained annotations of mask-audio pairs, making them impractical for scalability. To address this, we introduce unsupervised AVS, eliminating the need for such expensive annotation.
To tackle this more challenging problem, we propose an unsupervised learning method, named \textbf{M}odality \textbf{C}orrespondence \textbf{A}lignment (\modelone), which seamlessly integrates off-the-shelf foundation models like DINO, SAM, and ImageBind. This approach leverages their knowledge complementarity and optimizes their joint usage for multi-modality association. Initially, we estimate positive and negative image pairs in the feature space. For pixel-level association, we introduce an audio-visual adapter and a novel {pixel matching aggregation} strategy within the image-level contrastive learning framework. This allows for a flexible connection between object appearance and audio signal at the pixel level, with tolerance to imaging variations such as translation and rotation.
Extensive experiments on the AVSBench (single and multi-object splits) and AVSS datasets demonstrate that our \modelone{} outperforms strongly designed baseline methods and approaches supervised counterparts, particularly in complex scenarios with multiple auditory objects. Notably when comparing mIoU, \modelone{} achieves a substantial improvement over baselines in both the AVSBench (S4: \textbf{+17.24\%}; MS3: \textbf{+67.64\%}) and AVSS (\textbf{+19.23\%}) audio-visual segmentation challenges.

\end{abstract}    
\section{Introduction}
\label{sec:intro}
\vspace{-1em}
Audio-visual segmentation (AVS) accurately identifies pixel-level objects producing specific sounds in videos. Previous studies focused on audio-visual signal intersection using self-supervised learning methods~\cite{afouras2020self, rouditchenko2019self}, but face limitations in real-world applications like video editing and robotics. Supervised learning with pixel-level annotated video-audio pairs is intuitive~\cite{zhou2022audio}, but scaling annotation is challenging, as object segmentation demands over $100\times$ more effort than classifying or bounding boxes~\cite{zlateski2018importance}. In complex backgrounds, ground-truth segmentation labels may overlap or be ill-defined, requiring fine-grained boundaries for differentiation among surrounding objects.
\begin{figure}[t]
    \centering
    \includegraphics[width=0.86\textwidth]{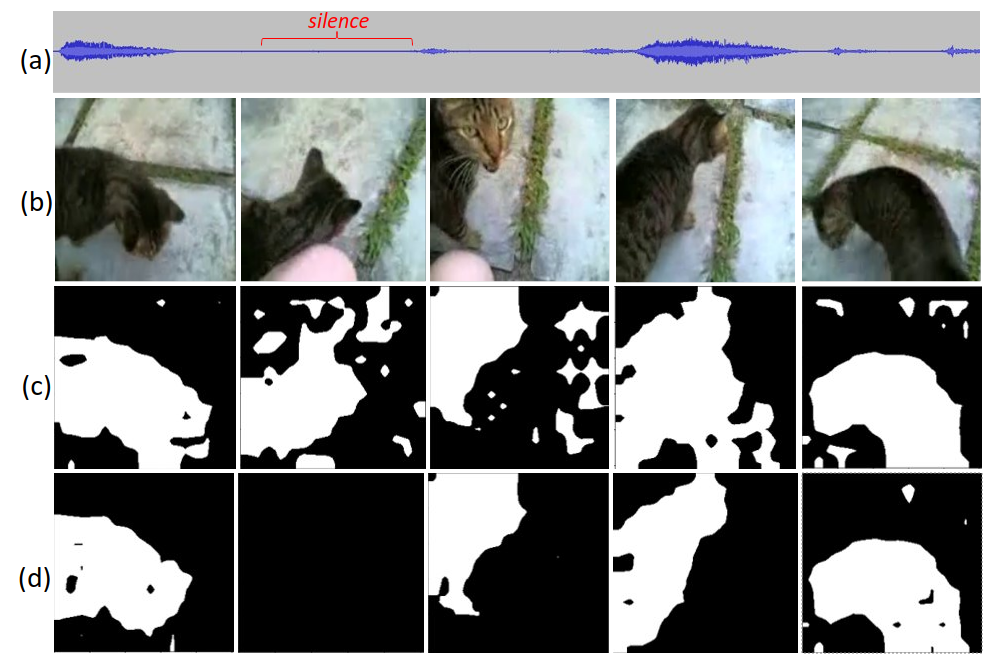}
    \caption{
    Emergence of coarse and noisy \textit{audio-pixel association}
    in ImageBind's \cite{girdhar2023imagebind} multimodal feature space. (a) Raw audio waveform, (b) input RGB, (c) using frozen ImageBind features, (d) \modelone{} (ours) -- generating finer masks and particularly only in the presence of sounding objects (see $2^{nd}$ frame, \modelone{} generates no mask due to silent audio).
    }
   \label{fig:fig_1}
\end{figure}
In this work, we advocate for addressing the practical and scalable  {\bf\em unsupervised AVS} problem, eliminating the need for exhaustive pixel annotation. This choice is influenced by notable advancements in purpose-generic image-level audio-visual representation learning, as demonstrated by ImageBind~\cite{girdhar2023imagebind}.
Empirically, we observe the emergence of coarse cross-modal associations between audio and image pixels, as depicted in Figure \ref{fig:fig_1}. However, despite this progress, the acquired knowledge remains insufficient for achieving an accurate unsupervised solution.

To address the mentioned limitation, we propose \textbf{M}odality \textbf{C}orrespondence \textbf{A}lignment (\modelone{}), an unsupervised audio-visual segmentation framework that efficiently integrates off-the-shelf foundation models using a minimal number of learnable parameters.
Our approach initiates with unsupervised pairing of training images, leveraging image-level visual features such as those provided by DINO \cite{dino}. This process augments the original audio-image pairing relationships by introducing additional positive and negative pairs at a coarse image level, thereby establishing the underlying {\em one-audio-to-multi-object-instances} mapping information. To establish precise audio-pixel associations, we introduce a novel audio-visual adapter and pixel matching aggregation within an audio-enhanced visual feature space. This strategy facilitates the correlation of audio signals and image pixels within a conventional image-level contrastive learning framework.
Specifically, we aggregate the similarity of pixel pairs across both the original and additionally paired images, effectively addressing imaging variations like translation and rotation, while allowing for the emergence of desired audio-pixel associations. To further refine object boundaries, we incorporate capabilities from open-world object detection and segmentation models (e.g., SAM \cite{kirillov2023segment}), aligning them with the derived audio-pixel associations. This integration ensures accurate delineation of object boundaries in the synchronized audio-visual context.

Our {\bf contributions} are summarized as follows:
\begin{itemize}
\setlength{\itemsep}{0pt}
\item We address the more challenging unsupervised Audio-Visual Segmentation (AVS) timingly, aiming to scale up cross-modal association capabilities for broader practical applications.
\item We propose a novel and efficient unsupervised framework, named \modelone{}, which systematically incorporates the off-the-shelf capabilities of pre-trained foundation models with a minimal number of additional parameters. This is facilitated primarily by our innovative 
pixel matching aggregation, enabling the automatic emergence of audio-pixel associations based on coarse image-level pairwise information.
\item We benchmark the performance of unsupervised AVS by introducing two robust baseline methods. Our extensive experiments demonstrate that \modelone{} significantly outperforms these baselines on both the AVSBench (S4: \textbf{+17.24\%}; MS3: \textbf{+67.64\%}) and AVSS (\textbf{+19.23\%}) datasets in terms of mIoU, thereby narrowing the performance gap with supervised AVS alternatives.
\end{itemize}
\section{Related Work}
\label{sec:related_work}
\vspace{-1em}
\textbf{Audio-Visual Semantic Segmentation}
Existing AVS methods use fully supervised models to identify audible visual pixels associated with a given audio signal \cite{zhou2022contrastive, mao2023contrastive, liu2023bavs, hao2023improving, shi2023cross, mo2023av}. These models are typically trained on thousands of manually annotated segmentation masks \cite{zhou2022audio}. However, their resource-intensive nature
for supervised training 
presents challenges for AVS systems, especially in large-scale applications with diverse scenes and multiple audible objects.
To address these challenges, we propose an unsupervised AVS approach in this work, eliminating the need for exhaustive audio-mask pairs.

\noindent\textbf{Self-Supervised Audio-Visual Feature Learning}
There have been various self-supervised approaches for learning audio-visual correspondence. Two primary methods, namely reconstruction by masked autoencoders (MAEs) \cite{georgescu2023audiovisual, nunez2023diffusion} and contrastive learning \cite{chen2021distilling, ma2020active, guzhov2022audioclip, wu2022wav2clip}, have been widely explored. Recent methodologies, such as MAVIL, integrate both MAE and contrastive learning to efficiently generate audio-visual representations for tasks like audio-video retrieval \cite{huang2022mavil}.
More recently, ImageBind, as presented in \cite{girdhar2023imagebind}, focuses on learning a joint embedding across six modalities, enhancing zero-shot capabilities for large-scale vision foundation models. These methods primarily address coarse relationships in a cross-modal manner. However, in the context of Audio-Visual Segmentation (AVS), our goal is to pinpoint object boundaries at the fine pixel level. Achieving this demands a more detailed and accurate understanding in the visual domain, and in this work, we address this challenge for the first time using an unsupervised approach.

\noindent\textbf{Unsupervised Image Semantic Segmentation}
Our challenge revolves around unsupervised image semantic segmentation, where the goal is to accurately isolate objects of interest within an image. Many approaches in the literature address this problem, primarily focusing on enhancing pre-trained self-supervised visual features.
One such method is SegSort \cite{segsort}, which employs an iterative process to refine visual features within a spherical embedding space. MaskContrast \cite{maskcontrast} takes a different approach by leveraging an off-the-shelf saliency model to generate binary masks for each image. The model then contrasts learned features within and across these saliency maps.
STEGO \cite{stego} addresses the problem by distilling the correspondence of visual features through self-correlation within the same image and cross-correlation across another similar image, along with complex post-processing steps.

It is worth noting that these methods specifically tackle mono-modality challenges and lack considerations for the cross-modal association between audio and pixel data—a unique aspect our work aims to address.
\section{Method}
\vspace{-1em}
Given a video sequence $\{I_t\}_{t=1}^{T} \in \mathcal{R}^{H \times W}$ comprising $T$ non-overlapping continuous image frames and an accompanying audio sequence $\{A_t\}_{t=1}^{T}$, our objective is to generate object segmentation masks $\{G_t\} \in \mathcal{R}^{H \times W}$ that correspond to the audio content. Here, $H$ and $W$ represent the height and width of each frame. These masks label individual pixels, highlighting the sound-producing object in $A_t$ within the frame $I_t$.
\begin{figure*}[t]
    \centering
    \includegraphics[width=0.94\textwidth]{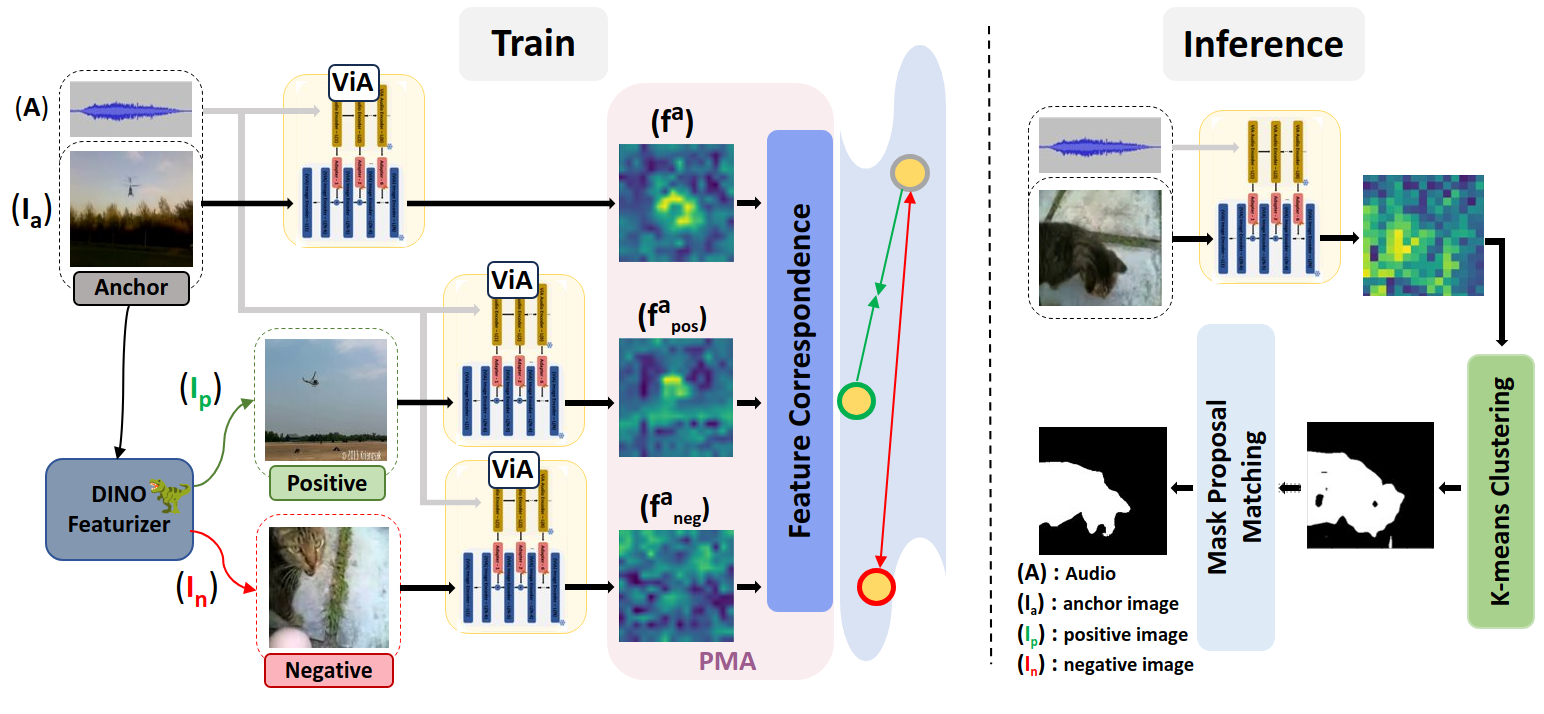}
    \caption{
    \textbf{Overview of our proposed \modelone{}}: (\textbf{Left}) 
    In training, we generate positive and negative images by utilizing DINO embeddings. The fusion of these images with the corresponding audio from the anchor image yields audio-enhanced image features. An efficient learning process by the proposed audio-visual adapter weights is facilitated through the establishment of a contrastive training objective, utilizing our pixel matching aggregation strategy.
    (\textbf{Right})
    In inference, we extract audio-enhanced image features and employ $k$-means clustering to form clusters. Optionally, we enhance object boundaries by matching the clustered feature map with mask proposals from a pre-trained SAM model. PMA refers to {\em pixel matching aggregation}. All vision-audio (ViA) model weights are shared and frozen.
    }
   \label{fig:comocoa_system}
\end{figure*}
In unsupervised AVS, detailed annotations for pixel-level masks are not available. To address this challenge, we propose an unsupervised approach named \modelone{}.
\vspace{-1.5em}
\subsection{\modelone{}}
\vspace{-0.6em}
As depicted in Figure \ref{fig:comocoa_system}, our \modelone{} leverages the existing capabilities of foundational models, including DINO \cite{dino}, ImageBind \cite{girdhar2023imagebind}, and SAM \cite{kirillov2023segment}, within a contrastive learning framework. Initially, we generate positive and negative image pairs using DINO embeddings. To seamlessly integrate these embeddings for unsupervised AVS, we propose an audio-visual adapter design with a minimal number of extra parameters. This design is integrated into a frozen multimodal foundation model, e.g., ImageBind, to produce audio-enhanced image features.

For accurate pixel segmentation association with the audio signal, we introduce a novel region matching aggregation strategy. This strategy facilitates the identification of object-level correspondences under the guidance of coarse image-level pairing supervision.
\noindent\paragraph{\bf Audio-Visual Adapter (AdaAV)}
The objective of our audio-visual adapter proposal is to integrate audio event-specific knowledge into frozen visual feature representations. This integration results in audio-enhanced image features that capture pixel-level correlations with audio signal. This forms the basis for exploring the inherent association between audio and visual objects.

The architecture for this design is illustrated in our supplementary. Although our subsequent discussion is grounded in the context of ImageBind \cite{girdhar2023imagebind}, it's important to note that our design is versatile, and integration with other multimodal Vision-Audio (ViA) models can be similarly achieved.

Each image frame, $I_t$, is individually inputted into ImageBind's image encoder. 
Within the encoder, $I_t$ undergoes decomposition into $n$ non-overlapping patches, which are then flattened into visual embeddings
$\mathbf{X}^{(0)}_v \in \mathbb{R}^{n \times d}$.
Simultaneously, $A_t$ is projected into Mel-spectrograms, patched, and fed into the ImageBind audio encoder, revealing audio embeddings denoted as $\mathbf{X}^{(0)}_a \in \mathbb{R}^{k \times d}$ (where
$0$ signifies layer-$0$, i.e., the input layer).
With $\mathbf{X}^{(\ell)}_a$ and $\mathbf{X}^{(\ell)}_v$ representing audio and visual inputs, respectively, at layer $l$,
both audio and image trunks employ a multi-headed attention (MHA) layer. The output from the MHA layer is then passed through a multi-layer perceptron (MLP) layer.

\begin{figure}[t]
    \centering
    \includegraphics[width=0.7\textwidth]{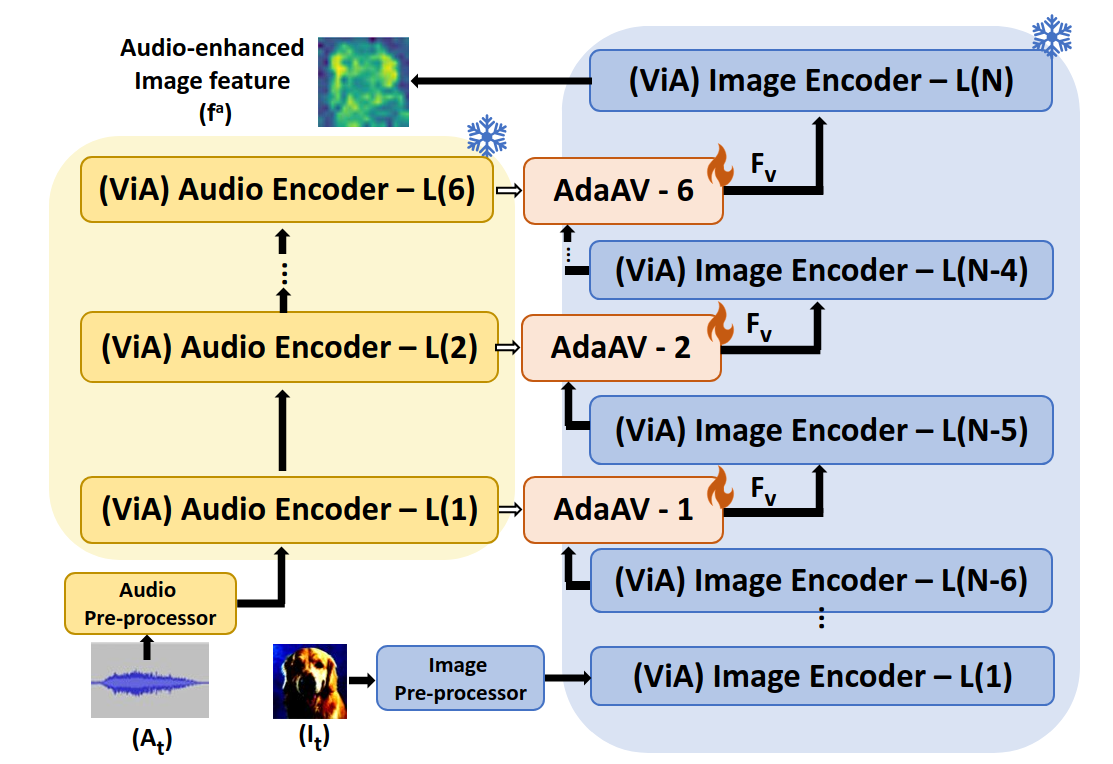}
    \caption{
    {\bf Fusing frozen ViA encoders using AdaAV}: 
    We augment frozen visual-audio (ViA) encoders using a lightweight audio-visual adapter, AdaAV, which, given an input image frame ($I_t$) and corresponding audio ($A_t$), generates an audio-enhanced image feature ($f^a$).}
   \label{fig:comoco_arch}
\end{figure}

Our audio-visual adapter module, AdaAV, is introduced between the intermediate layers of the image and audio trunks (please refer to Figure \ref{fig:comoco_arch}). 
For a given intermediate layer-$l$, AdaAV generates an audio-enhanced image feature representation, 
\begin{equation*}
\begin{aligned}
\mathbf{F}_v^{(\ell)} = \text{AdaAV}(\mathbf{X}_a^{(\ell)}, \mathbf{X}_v^{(\ell)})
\end{aligned}
\end{equation*}
The updated MHA and MLP operations in each layer of the image trunk are as follows:
\begin{equation*}
\begin{aligned}
\label{eq:mha_a2v}
\mathbf{Y}_v^{(\ell)} = \mathbf{X}^{(\ell)}_v + \mathrm{MHA} (\mathbf{X}^{(\ell)}_v) + \mathrm{AdaAV}(\mathbf{X}_a^{(\ell)}, \mathbf{X}_v^{(\ell)})\\
\mathbf{X}_v^{(\ell+1)} = \mathbf{Y}^{(\ell)}_v + \mathrm{MLP}(\mathbf{Y}^{(\ell)}_v) + \mathrm{AdaAV}(\mathbf{Y}_a^{(\ell)}, \mathbf{Y}_v^{(\ell)})
\end{aligned}
\end{equation*}

AdaAV employs a limited number ($m$) of latent audio tokens, denoted as $\mathbf{L}^{(l)}_a \in \mathbb{R}^{m \times d}$, to effectively integrate audio-specific knowledge into the visual representation. Here, the value of $m$ is significantly smaller than the overall number of audio tokens. The primary purpose of these latent tokens is to succinctly encapsulate information from the audio tokens, facilitating efficient information transfer into the intermediate layer visual representations.

Distinct sets of latent tokens are utilized at each layer. A cross-modal attention block is employed to compress all tokens from the audio modality into these latent tokens. Subsequently, another cross-modal attention block is utilized to fuse information between the compressed latent tokens of the audio modality and all tokens of the image modality.

In alignment with previous research on adapters~\cite{Adapter, perceiver, nips21_bottleneck}, a bottleneck module is incorporated. This module comprises a learnable down-projection layer, a non-linear activation function, and a learnable up-projection layer.

\subsubsection{Pixel matching aggregation (PMA)}
\label{sec:loss_function}
To achieve AVS, it is necessary to establish cross-modal associations at the pixel level. In pursuit of this goal, we delve into pixel-level correlations involved in audio-enhanced image features.
Drawing inspiration from the stereo matching literature \cite{prince2012computer, zisserman2004multiple}, we propose a {\em pixel matching strategy} that gauges the similarity between audio-enhanced image features generated at the pixel level. Formally, given two audio-enhanced feature tensors $f^a$ and $g^a$, a matching cost is computed for each spatial location in $f^a$, considering a set of pixels along the corresponding scanline of $g^a$. This cost signifies the matching likelihood of a pixel at a specific disparity in $f^a$ with a pixel in $g^a$. We further refine this process to identify optimal correspondences by pinpointing disparities with the lowest costs, signifying regions containing potential sounding source objects.

We compute a combined matching cost as:
\begin{equation}
\label{eq:pma}
    C = \lambda_{ssd} \cdot C_{ssd} + \lambda_{ncc} \cdot C_{ncc}
\end{equation}
where $\lambda_{ssd/ncc}$ is the weight hyper-parameter,
and the first term is the sum of squared differences (SSD) \cite{szeliski2022computer, trucco1998introductory}:
\begin{equation}
C_{ssd}(f^a, g^a) = \sum_{i, j} \left( f^{a}(i, j) - g^{a}(i, j) \right)^2
\label{eqn:correspondence_ssd}
\end{equation}
and the second term is the normalized cross-correlation (NCC) \cite{pratt2007digital}:
\begin{equation}
 C_{ncc}(f^a, g^a) = \frac{\sum_{i, j} \left( f^{a}(i, j) \cdot g^{a}(i, j) \right)}{\sqrt{\sum_{i, j} \left( f^{a}(i, j)^2 \right) \cdot \sum_{i, j} \left( g^{a}(i, j)^2 \right)}}
\label{eqn:correspondence_cc}
\end{equation}
Associating the pixels between two images in such a fashion permits a versatile matching of their objects, accommodating variations in imaging conditions such as object translation and rotation.
We posit that this dual-metrics cost brings about a complementary advantage, where SSD captures absolute intensity differences, and NCC considers both the similarity in shape and the scale of the signals. The positive impact of this fusion is empirically substantiated through experiments, as illustrated in Table \ref{tab:ablation_loss}. \swap{Given $f^a$, $g^a$ as the two audio-enhanced features whose similarity is to be computed, SSD captures absolute intensity by summing the squared discrepancies in pixel values between $f^a$ and $g^a$. NCC alternatively computes the similarity in the pattern or structure (shape) by normalizing the data, and hence less sensitive to variations in amplitude (scale).}
\subsubsection{Model training}
In our design, the AdaAV is the sole learnable module, while the audio and image encoder weights remain fixed. The learning of AdaAV weights is achieved through a contrastive training objective, which is outlined as follows:
\begin{align}
L_{c} = \max\left(0, C(f^{a}, f^{a}_{pos}) -  C(f^{a}, f^{a}_{neg}) + \alpha, 0\right)
\label{eqn:triplet_loss}
\end{align}
where $\alpha$ is the margin parameter
and $C$ is defined in Eq. \eqref{eq:pma}.
Positive ($f^{a}_{pos}$) and negative ($f^{a}_{neg}$) sets are formed by 
leveraging the pre-trained DINO \cite{dino} as an off-the-shelf image featurizer. 
Specifically, global image features are extracted using DINO embeddings through global average pooling. Subsequently, we create a lookup table containing each image's K-Nearest Neighbors (KNNs) based on cosine similarity in the DINO's feature space.
This augments the original audio-image pairs by introducing additional pairing information at the image level, providing the coarse {\em one-audio-to-multi-object-instances} association information. 
For any images $x$ and their corresponding random nearest neighbors, we form the set of positive images, denoted as $x_{pos}$. 
To constitute the set of negative images $x_{neg}$ relative to $x$, we randomly sample images by shuffling $x$ while ensuring that no image matches with itself or its top KNNs. The audio embedding $A$ corresponding to $x$ is then employed to augment both the positive set ($f^{a}_{pos}$) and negative set ($f^{a}_{neg}$), as illustrated in Figure \ref{fig:comocoa_system} for further details.
\subsubsection{Inference}
\label{sec:mask_proposals}
During the inference phase, when presented with an image and accompanying audio, we employ the AdaAV module to extract enhanced features from the audio-imbued image. Subsequently, we refine the learned pixel-level associations by applying a cosine distance-based k-means algorithm \cite{hartigan1979algorithm}. Additionally, we observe that overlaying the generated clusters with mask proposals derived from a pre-trained SAM model contributes to obtaining more precise object boundaries.

\begin{figure}[t]
    \centering
    \includegraphics[width=0.96\textwidth]{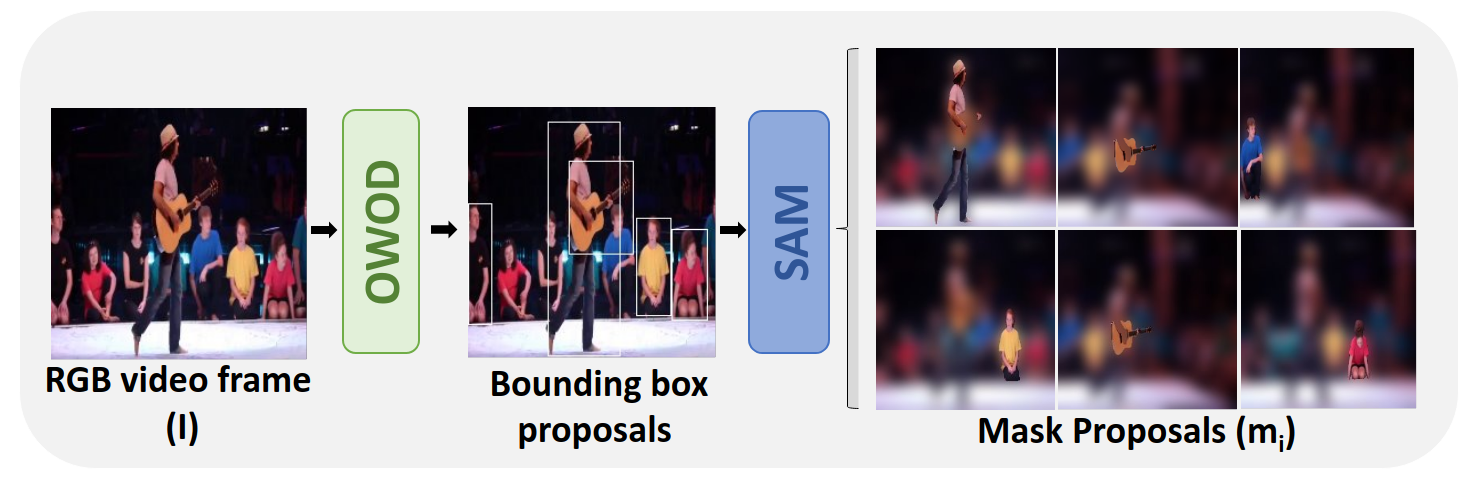}
    \caption{Generating mask proposals using a cascade of a pre-trained Open-world Object Detector (OWOD) \cite{maaz2022class} and Segment anything model (SAM) \cite{kirillov2023segment}. Mask proposals represent the existing visual objects in a given frame.}
   \label{fig:mask_proposals}
\end{figure}

To generate mask proposals, we initiate an open-world object discovery on the RGB image frame using a pre-trained Open World Object Detector (OWOD) model \cite{maaz2022class} (refer to Section \ref{sec:baselines} (II) for object discovery details). The bounding boxes produced by OWOD serve as input for the pre-trained SAM, resulting in proposal segmentation masks. These masks delineate visual objects within the given image frame, as depicted in Figure \ref{fig:mask_proposals}. By overlaying the mask generated from the audio-enhanced image feature onto these mask proposals, we effectively filter out potential auditory objects among the visual counterparts. \swap{Please note, the term overlay refers to choosing the proposal mask having IoU$>$0.5 with the mask generated from the fused ViA encoder (\ie from the audio-enhanced feature).}

It is important to note that while mask proposal matching is not a primary focus of our work, we conduct an ablative study in Section \ref{sec:sam_enh_ablation} to underscore the significance of our pixel matching strategy discussed earlier.

\def\modeltwo{{OWOD-BIND}}
\def\modelonebase{{AT-GDINO-SAM}}
\def\modelthree{{SAM-BIND}}

\section{Experiments}
\noindent
\textbf{Training Data} For training our AdaAV weights, we employ VGGSound dataset \cite{chen2020vggsound}, a large scale audio-visual dataset collected from YouTube. The dataset consists of over 200k clips for 300 different sound classes, spanning a large number of challenging acoustic environments and noise characteristics of real applications.

\noindent{\bf Benchmark settings}
To assess the proposed problem setting and method, we use the public AVSBench dataset \cite{zhou2022audio}. It contains videos from YouTube, split into 5 distinct clips. Each clip is paired with a spatial segmentation mask indicating the audible object. AVSBench includes two subsets: the \texttt{Semi-supervised Single Source Segmentation} (S4) and the fully supervised \texttt{Multiple Sound Source Segmentation} (MS3), differing in the number of audible objects. 

Additionally, we also report our scores on the \texttt{AVS-Semantic} (AVSS) dataset \cite{zhou2023audio}.
We evaluate our proposed method on the S4, MS3 and the AVSS test split without using any audio-mask pairs from the train and validation split (\ie, unsupervised AVS).
 
\noindent{\bf Performance metrics}
 We use average Intersection Over Union ($M_{IoU}$) and $F_{score}$ as metrics. A higher $M_{IOU}$ implies better region similarity, and an elevated $F_{score}$ indicates improved contour accuracy.
\subsection{Baselines}
\label{sec:baselines}
The existing supervised training framework for AVS demands resource-intensive audio-mask pairs annotations, not applicable for benchmarking unsupervised AVS.
To address the challenges, we present an alternative approach that eliminates the need for audio-mask pairs. This method leverages an untrained methodology informed by insights from sophisticated established foundational models.

Our proposal involves three meticulously designed integration mechanisms. They bootstrap foundation models related to unimodalities and associated tasks, such as segmentation, audio tagging, and visual object grounding.

\noindent{\bf(I) \modelonebase{}} In the first baseline, we adopt the Audio Spectrogram Transformer (AST) \cite{gong2021ast} to generate audio tags for the AVS task, utilizing its capacity to capture global context in audio sequences with multiple events. Trained on the large  Audioset dataset \cite{audioset}, AST detects diverse polyphonic audio events across 521 categories like animal sounds, instruments, and human activity.
Given an audio sequence, $A_t$, we pad it to a maximum of 960 msec and obtain its corresponding audio tags $\{AT_{i}\}_{i=1}^{C_a}$ using the pre-trained AST model. The tags are ranked based on their probability scores across 521 generic classes from the Audioset ontology. Relevant audio tags are filtered using an empirically determined threshold, $\tau_{AT}$ and forwarded to a pre-trained GroundingDINO model \cite{liu2023grounding}, generating bounding boxes in image frame $I_t$. These boxes serve as visual prompts for Segment Anything model (SAM) \cite{kirillov2023segment}, to produce producing binary masks.

\noindent{\bf(II) \modeltwo{}} We use a pre-trained Open World Object Detector (OWOD) \cite{maaz2022class} model to generate class-agnostic object proposals that are further processed by a segmentation pipeline. Unlike traditional object detectors which consider unfamiliar objects as background, OWOD models have been designed to handle unknown objects during training and inference. Given an image frame, $I_t$, we use OWOD to generate $C_v$  proposal bounding box proposals, $\{BB_{i}\}_{i=1}^{C_v}$. These are filtered by objectness score \cite{maaz2022class} using a threshold $\tau_{BB}$. To link boxes and acoustic cues, both modalities need a shared latent space embedding semantics. Towards this end, we utilize ImageBind's \cite{girdhar2023imagebind} latents, and extract image and audio embeddings which (post global average pooling) are used to rank the proposals by cosine similarity with the audio embedding. Bounding boxes above $\tau_{BIND}$ form the final mask.

\noindent{\bf(III) \modelthree{}}
Alternatively, instead of relying on the OWOD model to generate bounding box proposals, we can randomly position single-point input prompts in a grid across the image. From each point, SAM can predict multiple masks. These masks are refined and filtered for quality, employing non-maximal suppression (NMS) \cite{girshick2014rich} to remove duplicates. 

\subsection{Implementation Details}
\noindent\textbf{Baselines} We resize all image frames to 224 $\times$ 224. 
For all our evaluation results we use $\tau_{AT}$ as 0.5 for \modelonebase{} and $\tau_{BB}$ and $\tau_{BIND}$ as 0.5 and 0.7 respectively, for \modeltwo{}. For refining the mask outputs from \modelthree{}, we use IoU threshold of 0.5 for NMS.

\noindent\textbf{\modelone{}} For the frozen ViA we choose the publicly released ImageBind-Huge model with 6 and 12 transformer encoders in the audio and image trunks respectively. The resolution of the input images are 224 $\times$ 224. To optimize the model parameters, we employ the Adam \cite{loshchilov2017decoupled} optimizer with an initial learning rate of 1e-4 with cosine decay. We use $\lambda_{SSD}$=$\lambda_{NCC}$=1 and $\alpha$=0.3. 
We train the models for a maximum of 10000 iterations on a single NVIDIA RTX A5500 GPU. The batch size is set to 8.
For the mask proposal matching we consider only the proposal masks which have an IoU higher than 0.5 when compared with the mask generated from the fused ViA encoder (post $k$-means).
\subsection{Benchmark Results}
We first provide a quantitative comparison in terms of $M_{IoU}$ and $F_{score}$, followed by a more qualitative comparison in terms of the quality of the generated masks.

\noindent\textbf{(a) Quantitative Comparison:} We present our primary AVSBench results on the test set for both S4 and MS3, and the AVSS dataset in Table \ref{tab:main_table}.

\noindent\textbf{Comparing with salient object detection (SOD) and sound source localization (SSL) based approaches}, show a substantial gap between these and our \modelone{} in both S4 and MS3 subset. It is important to acknowledge that SOD operates solely on visual data without accounting for sound, prioritizing the dataset it was trained on. Consequently, it struggles with scenarios like MS3, where the auditory elements change while the visual components remain constant. On the other hand, SSL provides a more equitable comparison to our \modelone{}, as it incorporates audio signals to guide object segmentation. However, a significant disparity is apparent between SSL and both our \modelone{} and the proposed training-free baselines (\modelonebase{}, \modeltwo{}, \modelthree{}).

\noindent\textbf{Comparing with unsupervised baseline approaches},
we can observe that both, \modeltwo{} and \modelthree{}, outperform \modelonebase{} in terms of both $M_{IoU}$ and $F_{score}$ by a significant margin.
This is because AST itself achieves a mean average precision (mAP) of 0.485 when evaluated on the audio tagging task of Audioset, and hence the generated audio tags are prone to errors. Additionally, we believe despite AST's training data \ie, Audioset follows a generic ontology, many rare events (\eg ``lawn mover'', ``tabla'' etc.) are under-represented and hence are unable to cope with an open-set inference. 

We can observe that \modeltwo{} achieves an absolute improvement of 0.06 and 0.08
over \modelthree{} in terms of $M_{IoU}$ and $F_{score}$, respectively under the MS3 setting. Similar trend is observed under the S4 setting, with 0.16 improvement in both $M_{IoU}$ and $F_{score}$ respectively.

It can be clearly noted that \modelone{} surpasses all the baseline approaches by more than 10\% in terms of $M_{IoU}$ and more than 13\% in terms of the $F_{score}$ on both the S4 and MS3 settings. In contrast, to our best performing baseline \modeltwo, \modelone{} differs primarily in the latent space matching generated by the ImageBind encoders. This testifies our novel 
pixel-level association and matching strategy. 

\noindent\textbf{Comparing with supervised AVS approaches},
our unsupervised \modelone{} only falls short of the supervised AVS-Benchmark by 0.20 in terms of both $M_{IoU}$ and $F_{score}$ under the MS3 setting, where the latter benefits from the effective audio-visual contrastive loss design and trainable parameters.
This suggests both the promising potential of unsupervised AVS
and the impressive performance of our model design.
\begin{table}[t]
    \centering
    \caption{Performance comparison on the AVSBench 
    \texttt{test} split under the S4, MS3 and the AVSS setting. 
    Grayed: {\em supervised learning methods}.}
    \begin{tabular}{c|c|c|c|cc|cc|cc}
    
    \toprule
    & Approach & Mask-free & \# learnable & \multicolumn{2}{c}{S4}&\multicolumn{2}{c}{MS3}&\multicolumn{2}{c}{AVSS}\\
     & &training& params $\downarrow$&$\rm{M_{IoU}}\uparrow$&$\rm{F_{score}}\uparrow$&$\rm{M_{IoU}}\uparrow$&$\rm{F_{score}}\uparrow$&$\rm{M_{IoU}}\uparrow$&$\rm{F_{score}}\uparrow$ \\
    \midrule\midrule
    \rowcolor{lightgray}\parbox[t]{3mm}{\multirow{4}{*}{\rotatebox[origin=c]{90}{\tiny Supervised}}}&AV-SAM \cite{mo2023av} & \xmark & 1191M & $0.40$ & $0.56$ & - & - & - & -\\
   
     \rowcolor{lightgray}& 
     AVSBench \cite{zhou2022audio} & \xmark & 101M & $0.78$ & $0.87$ & $0.54$ & $0.64$ & 0.29 & 0.35\\
    
    \rowcolor{lightgray}&SAMA-AVS \cite{liu2023annotation} & \xmark & 75M & $0.81$ & $0.90$ & 0.63 & 0.69 & - & -\\
    
    \rowcolor{lightgray}&DG-SCT \cite{duan2024cross}& \xmark & 342M & $0.80$ & $0.89$ & 0.53 & 0.64 & - & -\\
    \midrule
    \parbox[t]{2mm}{\multirow{2}{*}{\rotatebox[origin=c]{90}{\tiny SOD}}}& iGAN \cite{mao2021transformer}& \cmark & 86M &$0.61$ & $0.77$ & $0.42$ & $0.54$ & - & -\\
    & LGVT \cite{zhang2021learning} & \cmark & 90M &$0.74$ & \bf{0.87} & $0.40$ & $0.59$ & - & -\\
    \midrule
    \parbox[t]{2mm}{\multirow{4}{*}{\rotatebox[origin=c]{90}{\tiny SSL}}}& LVS \cite{chen2021localizing}& \cmark & 20M &$0.37$ & $0.51$ & $0.29$ & $0.33$ & - & -\\
    & MSSL \cite{qian2020multiple} & \cmark & 21M &$0.44$ & $0.66$ & $0.26$ & $0.36$ & - & -\\
    & EZ-VSL \cite{mo2022localizing} & \cmark & 18M &$0.45$ & $0.68$ & $0.28$ & $0.34$ & - & -\\
    & Mix-Localize \cite{hu2022mix} & \cmark & 19M &$0.44$ & $0.69$ & $0.32$ & $0.36$ & - & -\\
    \midrule
    \parbox[t]{2mm}{\multirow{3}{*}{\rotatebox[origin=c]{90}{\tiny Baseline}}}& \modelonebase{} & \cmark & - & $0.38$ & $0.46$ & $0.25$ & $0.29$  & 0.24 & 0.25\\
    
    & \modelthree{} & \cmark & - &$0.42$ & $0.51$ & $0.28$ & $0.36$ & 0.24 & 0.26\\
    
    & \modeltwo{} & \cmark & - &0.58 & 0.67 & 0.34 & 0.44 & 0.26 & 0.29\\
    \midrule
   \rowcolor{cyan} \parbox[t]{3mm}{\multirow{1}{*}{\rotatebox[origin=c]{90}{\tiny Our}}}& {\bf \modelone{}} & \cmark & \bf{1.3M} & \bf{0.68} & 0.79 & \bf{0.57} & \bf{0.62} & \bf{0.31} & \bf{0.33}\\
    \bottomrule
    \end{tabular}
    \label{tab:main_table}
\end{table}

\begin{figure*}[t]
    \centering
    \includegraphics[width=0.99\textwidth]{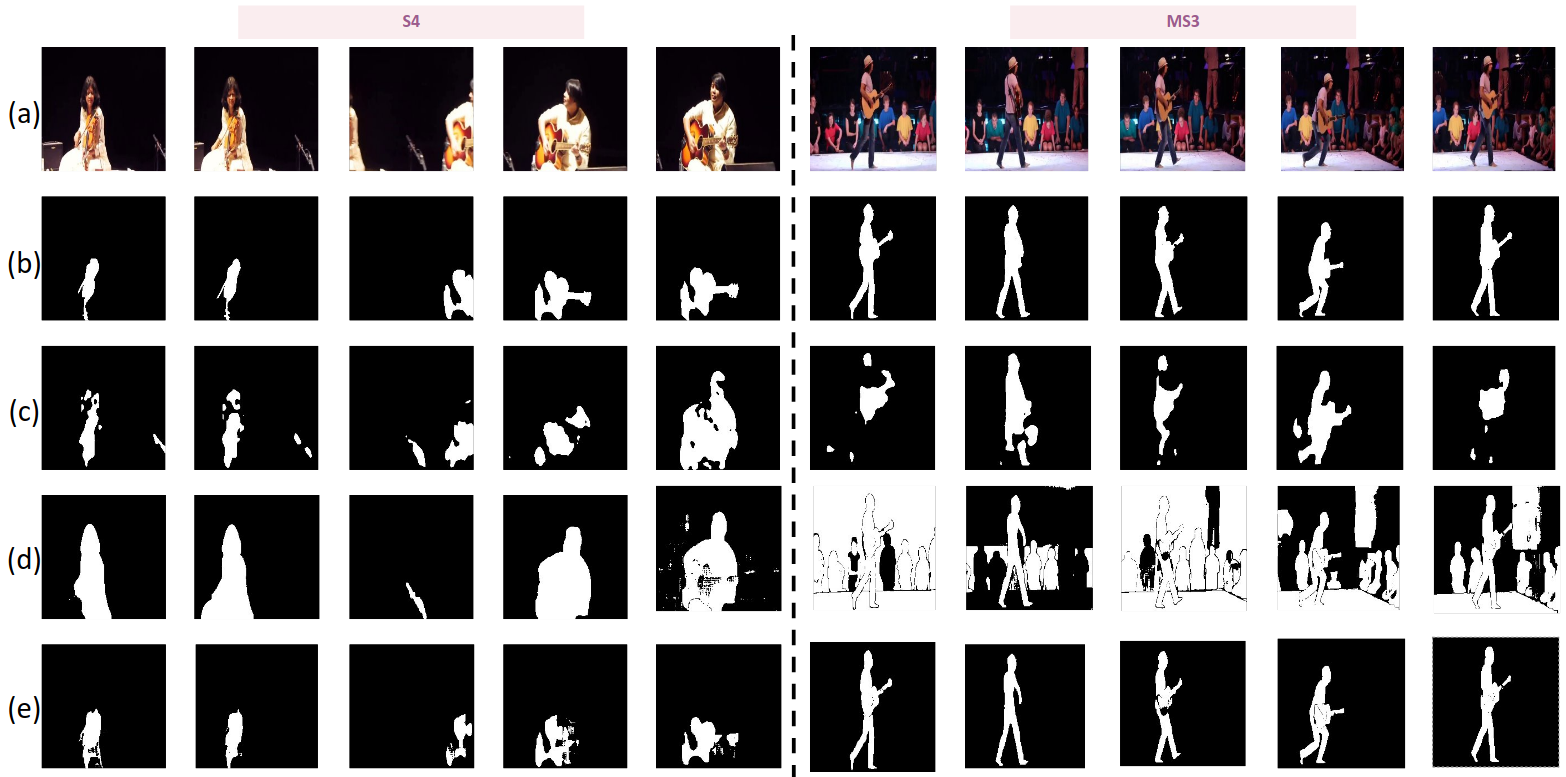}
    \caption{\textbf{Qualitative comparisons}(\textbf{left}: S4, \textbf{right}: MS3): (a) RGB frame, (b) ground truth mask, (c) AVSBench \cite{zhou2022audio}(supervised), (d) OWOD-BIND (baseline), (e) \modelone{} (ours). \modelone{} produces more precise segmentation of overlapping objects without utilizing any audio-visual masks during training.
    }
   \label{fig:main_plot}
\end{figure*}

\noindent\textbf{(b) Qualitative Comparison:} We show example segmentation from \modelone{} and the AVS supervised benchmark approach in Figure \ref{fig:main_plot}. We note that \modelone{} is significantly better at resolving fine-grained details within the images such as object boundaries (guitar in the left example and the guitarist in the right example). Although AVSBench uses a PVTv2 \cite{wang2021pyramid} architecture to output a high resolution segmentation, the network misses fine-grained details and generates discontinuous object segmentations. In contrast, \modelone{} despite not using any audio-mask pairs during training, can capture overlapping objects and fine details. 
It is evident that the baseline method, although is able to segment visual objects, but finds it difficult to filter out potential sounding objects among them.
In the circumstances when there are more than one objects of the same object category (\eg multiple humans) in the same frame, \modelone{}, unlike the baseline, is able to appropriately highlight the \texttt{"sounding human"}. Also, we provide qualitative comparison of \modelone{} and AVSBench on the AVSS split in the supplementary.
\subsection{Ablation Study}
\begin{figure*}[t]
    \centering
    \includegraphics[width=0.99\textwidth]{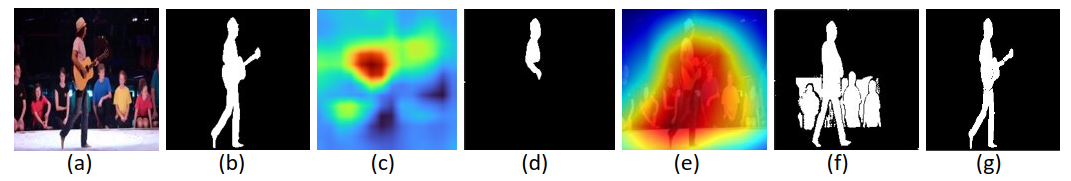}
    \vspace{-0.5em}
    \caption{\swap{\textbf{Comparing SSL methods equipped with MPM}: 
    (a) input frame, (b) ground truth mask, (c) Mix-Localize \cite{hu2022mix}, (d) Mix-Localize + MPM, (e) EZ-VSL \cite{mo2022localizing}, (f) EZ-VSL + MPM, (g) \modelone{} (ours) -- producing more precise segmentation of all overlapping sounding objects (Guitarist and guitar in this example).
    }}
   \label{fig:ssl_mpm}
   \vspace{-0.5em}
\end{figure*}

\vspace{-0.5em}
\noindent{\bf Effect of pixel matching aggregation:}
\label{sec:sam_enh_ablation}
As discussed in Section \ref{sec:mask_proposals}, we align the clustered predictions from \modelone{} with proposal masks, to obtain finer object segmentation boundaries. From Table \ref{tab:ablation_proposal_matching} it is evident that matching the proposals yields an overall improved segmentation performance, with maximum contribution of this performance owed to the audio-enhanced feature map learned by \modelone{}. For the scores in first row, we select random points as input to the SAM model\footnote{\tiny\url{github.com/facebookresearch/segment-anything/blob/main/notebooks/automatic_mask_generator_example}} and match all the segmentation maps with the closest ground truth in the MS3 test set. It can be noted that, this deprived of the mapping with \modelone{} generated enhanced feature, drops the performance significantly, with more than 0.2 drop compared to \modelone{}(row 3).
\swap{Additionally, we make it more concrete to particularly establish the higher contribution of PMA over MPM by combining existing SSL approaches with MPM. As can be seen in Table \ref{tab:ssl_mpm}, combining MPM with Mix-Localize \cite{hu2022mix} and EZ-VSL \cite{mo2022localizing} fails to surpass \modelone{} and even \modelone{} without MPM, due to the sparse nature of
SSL outputs. To further understand this, we pictorially highlight the under-segmentation (Fig. \ref{fig:ssl_mpm}(d)) and over-segmentation (Fig. \ref{fig:ssl_mpm}(f)) problems that occur when converting SSL output (heatmaps) to binary masks by employing MPM. Using this we reiterate the efficacy of pixel-level association learned by our proposed PMA strategy leading to fine-grained localization, while the MPM gives further (less significant than PMA) advantage by refining the object edges.}
\begin{table}[t]
\vspace{-2.5em}
  \centering
  \RawFloats
  \begin{minipage}[b]{0.45\textwidth}
    \centering
    \caption{Effect of pixel matching aggregation (PMA) and mask proposal matching (MPM) (Sec \ref{sec:mask_proposals}), on the performance of AVSBench MS3 \texttt{test} split.}
    \begin{tabular}{c|c|cc}
    \toprule
    PMA & MPM &\multicolumn{2}{c}{MS3}\\
&&$\rm{M_{IoU}}\uparrow$&$\rm{F_{score}}\uparrow$ \\
    \midrule\midrule
    \cmark & \xmark & 0.52 & 0.59\\
    \xmark & \cmark & 0.34 & 0.44\\
    \cmark & \cmark & \bf{0.57} & \bf{0.62}\\
    \bottomrule
    \end{tabular}
    \label{tab:ablation_proposal_matching}
  \end{minipage}
  \hfill
  \begin{minipage}[b]{0.45\textwidth}
    \centering
    \swap{
    \caption{Comparing \modelone{} and SSL methods equipped with mask proposal matching (MPM) on the AVSBench MS3 \texttt{test} split.}
    \begin{tabular}{c|c}
    \toprule
    Method & $M_{IoU}$ \\
    \midrule
   Mix-Loc\cite{hu2022mix}+MPM & 0.45\\
   EZ-VSL\cite{mo2022localizing}+MPM & 0.43\\
    \midrule
     {\bf MoCA} W/o MPM & {\bf 0.52}\\
     {\bf MoCA} & {\bf 0.57}\\
    \bottomrule
    \end{tabular}
    \label{tab:ssl_mpm}
    }
  \end{minipage}
  \vspace{-2.5em}
\end{table}

\noindent{\bf Loss function:} We ablate the impact of SSD and NCC components in calculating the loss function in Table \ref{tab:ablation_loss}. The results indicate that: 
(1) Using only NCC compared to only SSD results in a higher $M_{IoU}$ and $F_{score}$ (+6.55\%). This is primarily because SSD is sensitive to the intensity of the pixels when computing feature correspondences. 
(2) Combining SSD and NCC for computing the correlation leads to the best segmentation performance, confirming our design consideration. The NCC formulation for computing the correlation is invariant to the shape of the potential sounding region and hence agnostic to the scale of the sounding object among the anchor and positive images.
    
\begin{table}[t]
  \centering
  \RawFloats
  \begin{minipage}[b]{0.45\textwidth}
    \centering
    \caption{Effect of SSD and NCC on the performance of AVSBench S4 and MS3 \texttt{test} split. }
    \begin{tabular}{c|cc|cc}
    \toprule
    Loss & \multicolumn{2}{c}{S4}&\multicolumn{2}{c}{MS3}\\
     &$\rm{M_{IoU}}\uparrow$&$\rm{F_{score}}\uparrow$&$\rm{M_{IoU}}\uparrow$&$\rm{F_{score}}\uparrow$ \\
    \midrule\midrule
    SSD & 0.61 & 0.72 & 0.40 & 0.57\\
    NCC & 0.65 & 0.76 & 0.44 & 0.59\\
    SSD+NCC & \bf{0.68} & \bf{0.79} & \bf{0.57} & \bf{0.62}\\
    \bottomrule
    \end{tabular}
    \label{tab:ablation_loss}
  \end{minipage}
  \hfill
  \begin{minipage}[b]{0.45\textwidth}
    \centering
    \caption{Effect of the number of AdaAV blocks    on the performance of AVSBench S4 and MS3 \texttt{test} split.
    }
    \begin{tabular}{c|cc|cc}    
    \toprule
    \#(AdaAV) & \multicolumn{2}{c}{S4}&\multicolumn{2}{c}{MS3}\\     &$\rm{M_{IoU}}\uparrow$&$\rm{F_{score}}\uparrow$&$\rm{M_{IoU}}\uparrow$&$\rm{F_{score}}\uparrow$ \\
    \midrule\midrule
    2 & 0.60 & 0.69 & 0.39 & 0.47\\
    4 & 0.65 & 0.73 & 0.43 & 0.51\\
    6 & \bf{0.68} & \bf{0.79} & \bf{0.57} & \bf{0.62}\\
    \bottomrule
    \end{tabular}
    \label{tab:ablation_adaav_num}
  \end{minipage}
\end{table}

\noindent{\bf Number of adapter blocks:} Our proposed audio-visual adapters decide the amount of audio specific information that is added into the frozen image encoders. 
To examine the effect of this component design,
we vary the number of AdaAV blocks from 2 to 6 (the maximum number of adapters we can add since the ImageBind audio trunk consists of 6 transformer encoder blocks).
From Table \ref{tab:ablation_adaav_num}, we observe that increasing the number of adapter blocks improves both $M_{IoU}$ and $F_{score}$ by a huge margin albeit, non-uniformly with lesser positive impact when increasing from 4 blocks to 6. Using two AdaAV blocks yields the least performance indicating minimal audio-enhancement and hence poor audio-guided segmentation.

\noindent{\bf Placement of AdaAV blocks:} We experiment with the optimal placement of the AdaAV blocks relative to the ImageBind audio and image encoders blocks in Table. \ref{tab:ablation_ada_pos}. We fix number of AdaAV blocks to 6. Please note that the frozen ImageBind image encoder consists of 12 transformer encoder blocks. Notably, the 6 AdaAV blocks can be placed either along the first 6 (counting from the input layer) image encoder blocks, along the last 6 image encoder blocks, or interleaved such that every alternate encoder block in the image trunk is fused with one AdaAV block. Our observations are as follows: 
(1) Fusing every alternate encoder \ie interleaving, and fusing only blocks closer to the output \ie, last 6, yields almost similar performance with a max deviation of 0.01 in terms of $M_{IoU}$ on the MS3 setting. (2) Fusing blocks closer to the output yields a maximum of 3\% improvement in $M_{IoU}$ on the MS3 setting, indicating a higher adaptability is favoured closer to the latter layers.
\begin{table}[t]
    \centering
    \caption{Effect of positioning the AdaAV blocks w.r.t to the ImageBind image and audio trunk, on the performance of AVSBench S4 and MS3 \texttt{test} split. (k=6)}
    \begin{tabular}{c|cc|cc}    
    \toprule
    AdaAV & \multicolumn{2}{c}{S4}&\multicolumn{2}{c}{MS3}\\
Positioning&$\rm{M_{IoU}}\uparrow$&$\rm{F_{score}}\uparrow$&$\rm{M_{IoU}}\uparrow$&$\rm{F_{score}}\uparrow$ \\
    \midrule\midrule
    Interleaved & \bf{0.68} & 0.77 & 0.56 & 0.60\\
    first-k & 0.67 & 0.78 & 0.54 & 0.60\\
    last-k & \bf{0.68} & \bf{0.79} & \bf{0.57} & \bf{0.62}\\
    \bottomrule
    \end{tabular}
    \label{tab:ablation_ada_pos}
\end{table}
\section{Conclusion}
\vspace{-1em}
We introduce a more challenging unsupervised Audio-Visual Segmentation (AVS) problem, aiming to scaling its applicability.
For performance benchmarks, we show that harnessing self-supervised audio-visual models leads is effective.
Our proposed method, \textbf{M}odality \textbf{C}orrespondence \textbf{A}lignment (\modelone{}), features a novel 
pixel matching aggregation strategy for accurate pixel-audio associations.
Extensive experiments demonstrate that \modelone{} outperforms well-designed methods and closely approaches fully supervised counterparts. Notably, 
\modelone{} achieves substantial mIoU improvements in both the AVSBench (S4: \textbf{+17.24\%}; MS3: \textbf{+67.64\%}) and AVSS (\textbf{+19.23\%}) audio-visual segmentation settings.
\bibliographystyle{splncs04}
\bibliography{main}
\newpage
\setcounter{page}{1}

\section{Exploiting Audio-Visual Feature Correspondences from frozen ViA}
\begin{figure}[ht]
    \centering
    \includegraphics[width=0.72\textwidth]{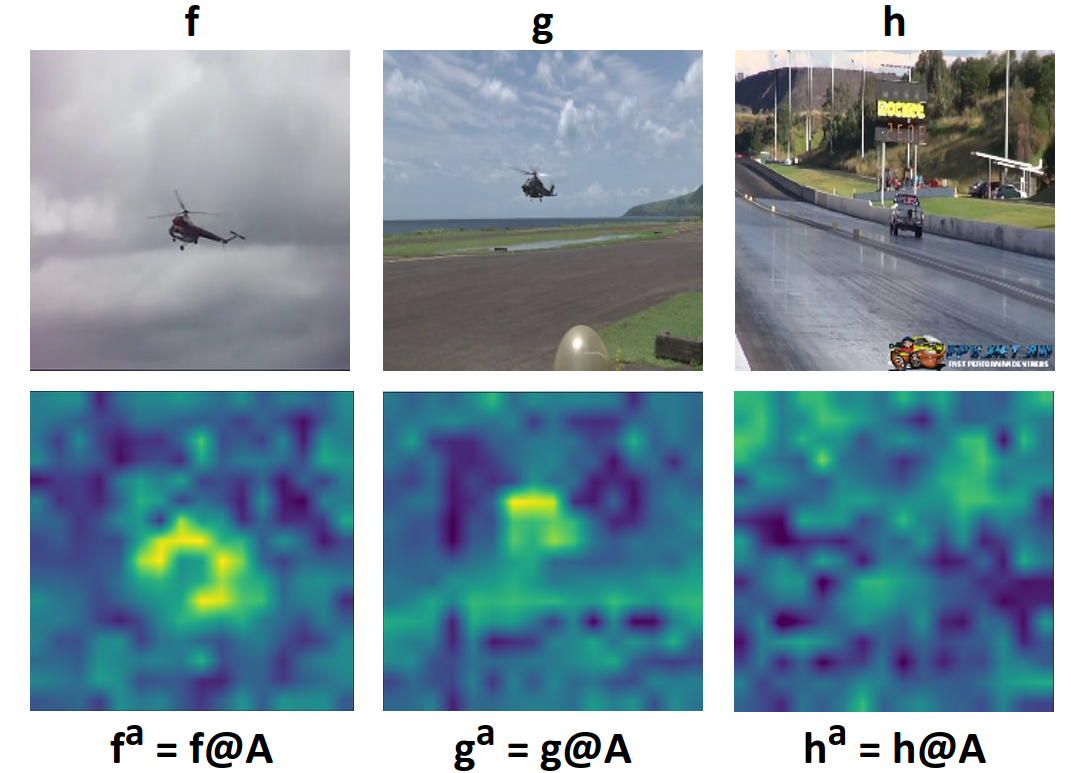}
    \caption{\textbf{Audio-Visual Feature Correspondence}: $A$ is the audio embedding of the sound of helicopter (\textit{AVSBench\-S4/train/6M\_JfKJxNVA.wav}). @ represents cosine similarity, computed at spatial (pixel) level.}
   \label{fig:correspondence}
\end{figure}

Recent progress in self-supervised audio-visual feature learning \cite{afouras2020self, alwassel2020self, cheng2020look, gong2022contrastive, georgescu2023audiovisual, girdhar2023imagebind} has yielded methods with powerful and semantically relevant features that improve a variety of downstream tasks. Though most works aim to generate vector representations for an image and a corresponding audio, there haven't been many works showing the semantic relevance among intermediate dense features. Towards this direction, we examine the correlation
between the dense feature maps, by fusing the audio and visual representations frozen ViA models. 
More formally, let $f\in \mathbb{R}^{EIJ}$
be the feature tensor for an image where $E$ represents the intermediate layer embedding dimension and $(I,J)$
represent the spatial dimensions. Additionally, let  $A\in \mathbb{R}^{E}$ be the audio embedding for the corresponding audio signal. We compute the audio-enhanced image feature tensor by taking the cosine similarity of $A$ with every spatial (or pixel) level feature $f_{ij} \forall i \in I$ and $ \forall j \in J$ in the image tensor $f$, \ie $f^{a}_{ij} = cosine\_sim(f_{ij}, A)$. Notably, we notice that for two image features $f$ and $g$, the introduction of the same audio embedding $A$ results in correlated tensors $f^a$ and $g^a$, provided that images corresponding to $f$ and $g$, consist a potential sounding object described by the audio representation in $A$.
For instance, Figure \ref{fig:correspondence} shows how the audio-enhanced features for three different images when enhanced with the same audio are in correspondence with relevant semantic areas particularly around the potential sounding object. It is evident that, in contrast to $f^a$ and $g_a$, $h_a$ does not effectively emphasize a distinct potential sounding area that might serve as the likely source for the sound of a helicopter, especially considering the absence of a helicopter in the image.

\section{Related tasks to AVS - SSL and SOD}
Sound source localization (SSL), saliency object detection (SOD), and audio-visual segmentation are all related tasks that involve processing audio and visual information to understand and analyze a scene. 
AVS aims to identify and separate audio and visual elements within a scene, such as identifying which sound sources correspond to which visual objects or events. SSL is limited to identifying the location of different sound sources in the scene, wherein the location need not be particularly as fine-grained as the output of an AVS task. SOD models are trained with the objective to identify visually salient objects or regions in a visual scene. Tt is important to be noted that there is no audio cue involved in SOD, or in other words, the salient object may or may not be a sounding object. To conclude, while all three tasks, localize the objects in a visual scene, SSL is a more closely related task to AVS, than SOD. This is also testified from our observations made in Section 4.3 (main paper) and Table 1 (main paper).

\subsection{Additional comparison with Sound Source Localization (SSL) methods}
In Table \ref{tab:ablation_ssl_comparison}, we report the results of two state-of-the-art SSL methods, \ie, LVS \cite{chen2021localizing} and MSSL \cite{qian2020multiple} on the AVSBench MS3 test set. LVS uses the background and the most confident regions of sounding objects to design a contrastive loss, followed by obtaining the localization map by computing the audio-visual similarity. MSSL is a two-stage method for multiple SSL and the localization map is obtained by Grad-CAM \cite{selvaraju2017grad}. 
\begin{table}[ht]
    \centering
    \caption{Comparing \modelone{} with SoTA SSL approaches on AVSBench MS3 \texttt{test} split. }
    \begin{tabular}{c|cc}
    
    \toprule
    Method &\multicolumn{2}{c}{MS3}\\
     &$\rm{M_{IoU}}\uparrow$&$\rm{F_{score}}\uparrow$\\
    \midrule\midrule
    LVS \cite{chen2021localizing} & 0.29 & 0.33\\
    MSSL \cite{qian2020multiple} & 0.26 & 0.36\\
    \modelone{}(w\ o MPM) & 0.52 & 0.59\\
    \modelone{} & \bf{0.57} & \bf{0.62}\\
    \bottomrule
    \end{tabular}
    \label{tab:ablation_ssl_comparison}
\end{table}
There is a substantial gap between the results of SSL methods and our \modelone{}. We hypothesize this is primarily because, unlike \modelone{}, the SOTA SSL methods lack a pixel-level aggregation strategy. This is even straightforward from our comparison in row 3, \ie, \modelone{} without the mask proposal matching (MPM), which essentially rules out any contribution of the mask matching from the pre-trained SAM model. This strengthens our claim regarding the importance of the pixel matching aggregation (PMA) in \modelone{}.

\subsection{How does \modelone{} perform in a SSL test setup?}
\swap{Following the test setup adopted in \cite{hu2022mix}, we present our scores if \modelone{} 
 was to be utilized for the SSL task. Similar to \cite{hu2022mix}, we present our scores on the VGGSound-Single dataset \cite{chen2020vggsound} (Table \ref{tab:moca_on_ssl}). Owing to the fine-grained localization, MoCA surpasses SSL methods in both, IoU@0.5 as well as AUC.
}
\begin{table}[ht]
    \centering
    \begin{tabular}{c|c|c}
    \toprule
    Method & $IoU@0.5$ & AUC\\
    \midrule
   Mix-Loc\cite{hu2022mix}&36.3&38.9\\
   EZ-VSL\cite{mo2022localizing}&38.9&39.5\\
     \bf{MoCA} &\bf{40.6}&\bf{43.5}\\
    \bottomrule
    \end{tabular}
    \caption{\textcolor{black}{SSL on VGGSound-Single dataset \cite{chen2020vggsound}}}
    \label{tab:moca_on_ssl}
\end{table}

\section{Qualitative analysis on the AVSS (semantic segmentation) split.}
\begin{figure}[ht]
    \vspace{-1em}
    \centering
    \includegraphics[width=0.98\linewidth]{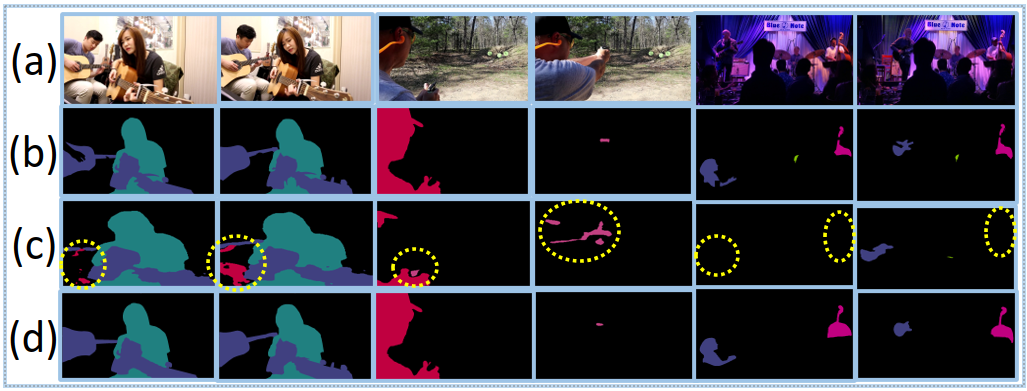}
    \caption{Qualitatively comparing \modelone{} with AVSBench (supervised) method, on the AVSS split.}
   \label{fig:avss_vis}
\end{figure}
\swap{
For obtaining semantic labels for AVSS using \modelone{}, we utilize our baseline, SAM-BIND. The use of MPM stage is same as the inference of S4 and MS3 setups, but, we distinguish between the mapped MPM masks if their IoU$<$0.3. 
As shown depicted in Fig. \ref{fig:avss_vis} RGB frame, (b) GT, (c) AVSBench-Supervised\cite{zhou2023audio}, (d) MoCA.
We encircle the regions where \cite{zhou2023audio} over-segments or under-segments objects. 
In frames$\{1,2\}$, compared to MoCA, \cite{zhou2023audio} over-segments the silent person in the background. In frames$\{3,4\}$, \cite{zhou2023audio} fails to segment the gun owing to its small size. Again, in frames$\{5,6\}$, \cite{zhou2023audio} fails to segment both, the lady in the audience (bottom-left) and the instrument (top-right). Overall, the efficacy of our pixel matching aggregation (PMA), weighing both the audio and visual modality appropriately, is shown in the last row.
}

\section{Ablation study on varying the margin in triplet loss}
The margin parameter defines a minimum difference that should be maintained between the distances of the anchor-positive pair and the anchor-negative pair. We ablate this hyper-parameter and report the performance of \modelone{} on the AVSBench MS3 test split in Table \ref{tab:ablation_triplet_margin}. We observe that optimal performance in terms of both the metrics, is obtained with a margin of 0.3, we argue that choosing a smaller margin makes the loss function that is too easy to satisfy and the model might converge to a solution where the positive and negative pairs are not well separated. Moreover, increasing the margin further up to 0.5, enforces a stricter condition on the distances, making the optimization problem more challenging.

\begin{table}[ht]
    \centering
    \caption{Effect of varying the margin in triplet loss, on the performance of AVSBench MS3 \texttt{test} split.}
    \begin{tabular}{c|cc}
    \toprule
    Margin &\multicolumn{2}{c}{MS3}\\
($\alpha$)&$\rm{M_{IoU}}\uparrow$&$\rm{F_{score}}\uparrow$ \\
    \midrule\midrule
    0.1 & 0.54 & 0.56\\
    0.3 & \bf{0.57} & \bf{0.62}\\
    0.5 & 0.56 & 0.58\\
    \bottomrule
    \end{tabular}
    \label{tab:ablation_triplet_margin}
\end{table}

\section{Baseline approaches}
\begin{figure*}[t]
    \centering
    \includegraphics[width=0.96\textwidth]{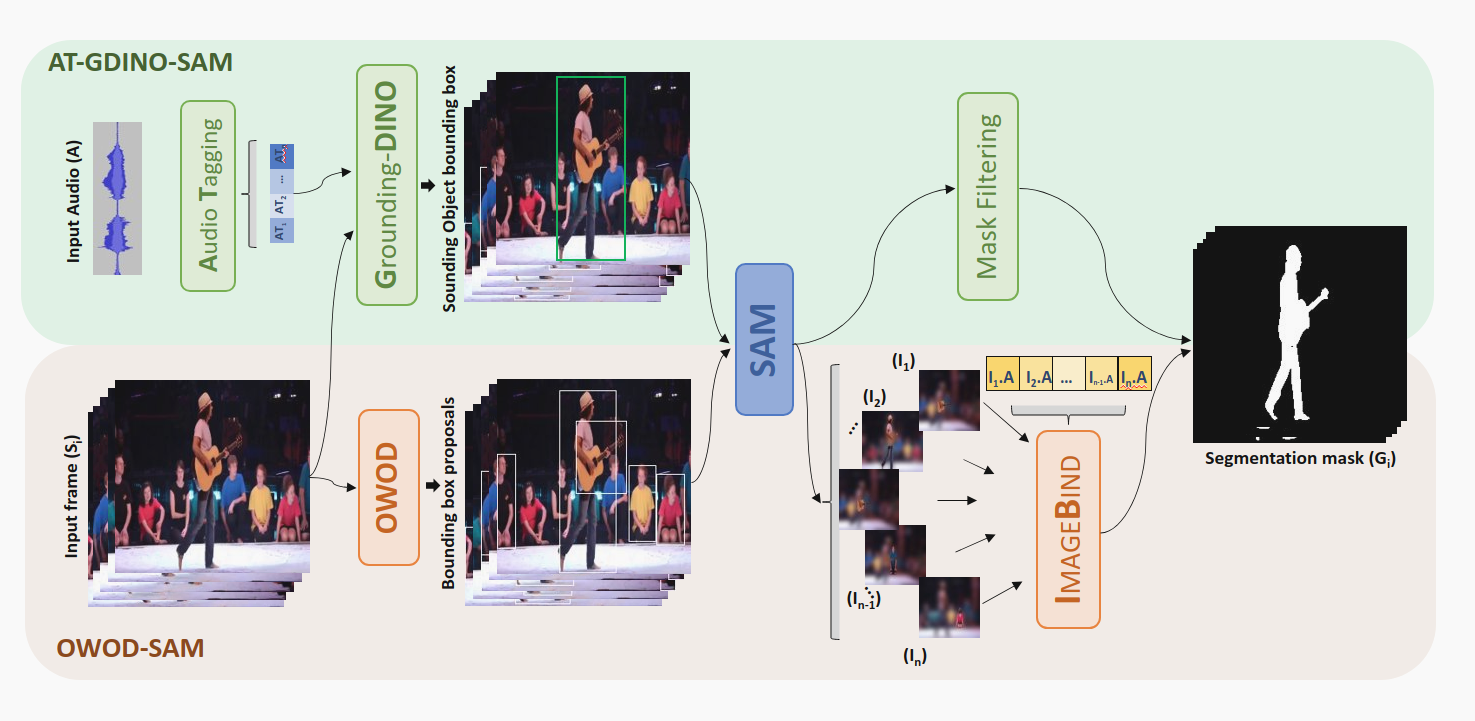}
    \caption{\textbf{Overview of baseline models}: Bounding box proposals are generated using a cascade of two pre-trained models: Audio Tagging and GroundingDINO. Favourable boxes are passed as input to SAM to yield segmentation masks.}
   \label{fig:baseline_models}
\end{figure*}

For tackling this newly proposed problem, Unsupervised AVS, towards establishing strong baselines, we formulate a novel {\bf\em Cross-Modality Semantic Filtering} (CMSF) approach to accurately associate the underlying audio-mask pairs by leveraging the off-the-shelf multi-modal foundation models (\eg, detection 
\cite{gong2021ast}, open-world segmentation \cite{kirillov2023segment} and multi-modal alignment \cite{girdhar2023imagebind}).
Guiding the proposal generation by either audio or visual cues, we design two training-free variants: \modelonebase{} and \modeltwo{}, respectively. Figure \ref{fig:baseline_models} shows a pictorial representation of our baseline approaches. 

\section{Ablation study on varying the IOU match for mask proposal matching (MPM)}
In the MPM stage, we match the audio-enhanced clustered features (generated as an outcome of the PMA) with the proposal masks from the pre-trained SAM model. The proposal masks represent all visual objects present in the scene, whereas the PMA clusters represent the sounding object. In Table \ref{tab:ablation_mpm_iou}, we vary the IOU threshold that is set to choose a proposal mask as the part of the final AVS output. In simpler words, when matching the PMA cluster with the proposal mask, if the IOU is above a certain threshold only then we consider the proposal mask as a part of the final AVS output.  A lower threshold yields "white" masks, since almost every proposal object mask is considered into the final output. Conversely, a higher threshold, yields "black" masks, since almost no object is shown in the final output due to poor match with the PMA clusters. 
We observe that generally a higher value of threshold is preferred.

\begin{table}[ht]
    \centering
    \caption{Effect of varying the IOU match for mask proposal matching (MPM), on the performance of AVSBench MS3 \texttt{test} split.}
    \begin{tabular}{c|cc}
    \toprule
    IOU &\multicolumn{2}{c}{MS3}\\
threshold &$\rm{M_{IoU}}\uparrow$&$\rm{F_{score}}\uparrow$ \\
    \midrule\midrule
    0.1 & 0.35 & 0.39\\
    0.3 & 0.48 & 0.48\\
    0.5 & \bf{0.57} & \bf{0.62}\\
    0.7 & 0.52 & 0.54\\
    0.9 & 0.54 & 0.54\\
    \bottomrule
    \end{tabular}
    \label{tab:ablation_mpm_iou}
\end{table}


\end{document}